\documentclass[runningheads]{llncs}
\usepackage[T1]{fontenc}
\usepackage{graphicx,verbatim}
\usepackage{amsmath}
\usepackage{amsfonts}
\usepackage{microtype}
\usepackage{tabularx,booktabs}
\usepackage{multirow}
\usepackage{placeins}
\usepackage{subcaption}
\usepackage{cite}
\newcolumntype{C}{>{\centering\arraybackslash}X} 
\usepackage{lipsum} 
\usepackage{pifont}
\begin{document}
\title{PET-Adapter: Test-Time Domain Adaptation for Full and Limited-Angle PET Image Reconstruction}
\titlerunning\space{PET-Adapter}
\author{Rüveyda Yilmaz\inst{1} \and
Yuli Wu\inst{2} \and
Johannes Stegmaier\inst{2} \and
Volkmar Schulz\inst{1}}

\authorrunning{Yilmaz et al.}

\institute{
Institute of Imaging and Computer Vision, RWTH Aachen University, Germany \and
Machine Learning for Medical Data, Heinrich Heine University Düsseldorf, Germany \\ Email:\email{rueveyda.yilmaz@lfb.rwth-aachen.de}
}

\maketitle              
\begin{abstract}
Positron Emission Tomography (PET) image reconstruction is inherently challenged by Poisson noise and physical degradation factors, which are further exacerbated in limited-angle acquisitions.
While deep learning methods demonstrate promising performance, their generalization to unseen clinical data distributions remains limited without extensive retraining.
We propose PET-Adapter, a test-time domain adaptation framework for generative PET reconstruction models pretrained solely on phantom data.
Our method enables adaptation to clinical datasets with varying anatomies, tracers, and scanner configurations without requiring paired ground truth.
PET-Adapter introduces layer-wise low-rank anatomical conditioning during adaptation and Ordered Subset Expectation Maximization-based warm-starting that initializes the generation from physics-informed reconstructions, reducing diffusion steps from 50 to 2 without compromising quality.
Experiments across multiple clinical datasets demonstrate superior 3D reconstruction performance in both full-angle and limited-angle settings, highlighting the clinical feasibility and computational efficiency of the proposed approach.
\end{abstract}
\section{Introduction}

Positron Emission Tomography (PET) is a powerful medical imaging modality widely used for the diagnosis, treatment planning, and therapeutic monitoring of diseases such as cancer, epilepsy, and Alzheimer’s disease \cite{hashimoto2024deep,zhang2022aerobic,cherry2006pet}. 
Despite its clinical value, PET image reconstruction remains challenging due to Poisson noise, Compton scattering, random coincidences, and detector gaps \cite{cherry2006pet}.
Conventional iterative reconstruction methods \cite{osem,bowsher2004utilizing} partially mitigate these effects; however, they are limited in their ability to produce high-quality images and are increasingly being challenged by Deep Learning (DL) approaches \cite{webber2024diffusion}.

DL-based post-processing methods can be applied in the image domain to denoise reconstructions from conventional algorithms \cite{schaefferkoetter2020convolutional,liu2019higher}. 
Although effective, these approaches are inherently constrained by the quality of the initial reconstruction and do not leverage raw measurement data.
In contrast, end-to-end models are trained on measurement–ground truth (GT) pairs and act as standalone reconstructors \cite{whiteley2020directpet,liu2022deep}. 
While these methods often produce sharper images, they may hallucinate structures that are inconsistent with the measured data \cite{hashimoto2024deep}.
To address this, DL-based iterative methods integrate neural networks as regularizers or learned priors within classical optimization frameworks (e.g., ADMM or unrolled FBSEM), enforcing stronger data consistency \cite{webber2024diffusion,singh2023score,9123435}.
To reduce the reliance on large, paired training datasets, recent work has shifted toward unsupervised methods.
Deep Image Prior (DIP)-based methods have been adapted to PET sinogram and list-mode data, demonstrating improved preservation of high-resolution information \cite{hashimoto2023fully,10025780}. 
More recently, generative modeling has advanced toward diffusion models with PET-specific adaptations and joint PET-magnetic resonance (MR) image reconstruction \cite{singh2023score,xie2024joint}.

Beyond standard reconstruction challenges, limited-angle (LA) PET introduces additional difficulties due to incomplete measurements from dual-panel geometries or partial-ring scanners \cite{raj2024recovery,makkar2024partial}.
Prior approaches to LA artifact mitigation include supervised U-Nets to map incomplete data to full-ring equivalents \cite{raj2024recovery}, conditional diffusion models for sinogram inpainting \cite{11287858}, or implicit neural representations \cite{makkar2024partial}. 
However, these approaches for both full-angle (FA) and LA PET image reconstruction often exhibit limited generalization to previously unseen data distributions, thereby constraining their practical applicability.
As a result, there remains a need for methods capable of adapting to variations in scanners, tracers, and anatomical characteristics without requiring extensive retraining on target data with paired reference GT images.
Hashimoto et al. \cite{hashimoto2026pet} and Webber et al. \cite{webber2025steerable} address this problem by adapting a Deep Diffusion Image Prior and a Pretrained Score-Based Generative Model, respectively, on a per-scan basis.
We introduce PET-Adapter, a dataset-wide adaptation strategy for Test-Time Domain Adaptation (TTDA) that generalizes to unseen clinical datasets using a model pretrained solely on phantom images.
Our main contributions are as follows: (1) We present an unsupervised adaptation framework with a tailored loss function to transfer generative PET reconstruction models trained on phantom data to clinical datasets with varying anatomies, pathologies, and tracers, thereby improving 3D reconstruction performance in both FA and LA settings. (2) We propose a strategy to incorporate MR images into unconditional pretrained models during adaptation via layer-wise conditioning \cite{perez2018film,stracke2024ctrloralter}, providing anatomical guidance. (3) We introduce a warm-starting strategy via diffusion inversion \cite{kim2022diffusionclip,songdenoising,yilmaz2025cellstyle} to initialize the inference with reconstructions from the Ordered Subset Expectation Maximization (OSEM) method \cite{osem}, reducing the diffusion sampling steps from 50 to 2 while maintaining quality.
\section{Method}
\textbf{PET image reconstruction} is an ill-posed inverse problem, primarily due to the ill-conditioned nature of the system matrix and the stochastic Poisson noise inherent to radioactive decay.
This condition is further exacerbated by physical degrading factors such as photon attenuation, scattering, and random coincidences \cite{singh2023score}.
The statistical model of the acquisition process is defined as:
\begin{equation}
\mathbf{y} \sim  \operatorname{Poisson}(\mathbf{A} \mathbf{x}+\mathbf{b}),
\end{equation}
where $\mathbf{y} \in \mathbb{N}^{m}$ represents the measured projection data (prompts).
$\mathbf{A} \in \mathbb{R}^{m \times n}$ is the system matrix (or forward projector), modeling the probability that an emission from voxel $j$, where $j \in \{0,1,...,n-1\}$, is detected by the detector pair $i$, where $i \in \{0,1,...,m-1\}$.
$\mathbf{x} \in \mathbb{R}^{n}_{\geq 0}$ denotes the unknown radiotracer activity distribution, and $\mathbf{b}\in \mathbb{R}^{m}$ accounts for the expected additive background events; namely, scattered and random coincidences.
PET image reconstruction aims to estimate the activity distribution $\hat{\mathbf{x}}$ from the noisy measurements $\mathbf{y}$.

\noindent\textbf{Score-based generative models} can generate samples by learning to reverse a stochastic noising process described by Stochastic Differential Equations (SDEs) \cite{songscore}.
While the standard reverse process requires many iterative steps, the Denoising Diffusion Implicit Model (DDIM) formulation allows for accelerated sampling \cite{songdenoising}.
\begin{figure}[tbp]
\centering
\includegraphics[width=1.0\textwidth]{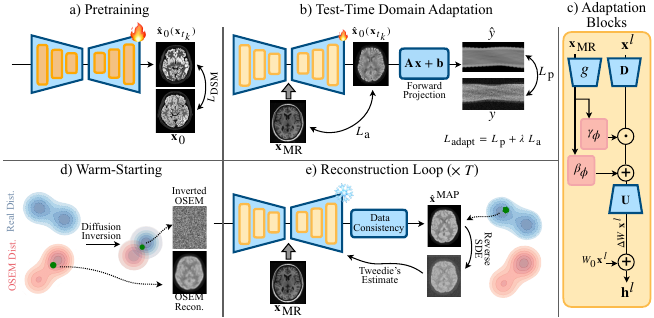}
\caption{The architecture of PET-Adapter. (a) Pretraining on phantom data is followed by (b) test-time domain adaptation on clinical target domains via (c)~layer-wise adaptation blocks. The final reconstruction is initialized with (d)~warm-starting (WS) and (e) optimized through $T$ alternating steps of prior refinement and data-consistency updates.}
\label{fig:main}
\end{figure}%
A generalized update step at timestep $t_k \in \left(0,1\right] $, which can interpolate between deterministic DDIM sampling and stochastic variance-preserving SDE, is formulated as:
\begin{equation}
\mathbf{x}_{t_{k-1}}=\gamma_{t_{k-1}}\hat{\mathbf{x}}_0\left(\mathbf{x}_{t_k}\right)-v_{t_k}\sqrt{v_{t_{k-1}}^2-\eta_{t_k}^2} s_\theta\left(\mathbf{x}_{t_k}, t_k\right)+\eta_{t_k} \mathbf{z} \text { with } \mathbf{z} \sim \mathcal{N}(\mathbf{0}, \mathbf{I}),
\end{equation}
where $\hat{\mathbf{x}}_0\left(\mathbf{x}_{t_k}\right):=(\mathbf{x}_{t_k}+v_{t_k}^2 s_\theta\left(\mathbf{x}_{t_k}, t_k\right))/{\gamma_{t_k}}$ denotes Tweedie’s estimate of the clean image derived from the noisy state $\mathbf{x}_{t_k}$, and $s_\theta$ is the learned score function.
The coefficients $\gamma_{t}$, $v_{t}$, and $\eta_{t}$ are time-dependent scalars that define the signal schedule (mean and variance) and the degree of stochasticity, respectively \cite{singh2023score}.
$s_\theta$ is trained with the standard denoising score matching (DSM) loss $L_\mathrm{DSM}$~\cite{vincent2011connection}.
To adapt this sampling approach for PET image reconstruction, Singh et al. \cite{singh2023score} propose a framework with a Maximum A Posteriori (MAP) objective that enforces data consistency, producing $\hat{\mathbf{x}}^{\mathrm{MAP}}$.
This is achieved by splitting the update into a prior term (U-Net-based diffusion) and a likelihood term.
Specifically, the intermediate estimate $\hat{\mathbf{x}}_0$ is refined by maximizing the penalized Poisson log-likelihood (MAP objective) using preconditioned gradient ascent with the standard expectation maximization preconditioner.
For our experiments in Section~\ref{section:experiments}, we employ this approach as the phantom-pretrained model (see Fig.~\ref{fig:main}a) to showcase the adaptation capabilities of PET-Adapter on novel datasets.
\begin{table*}[tbp]
\centering
\footnotesize
\newcolumntype{Y}{>{\centering\arraybackslash}X}
\setlength{\tabcolsep}{3pt} 

\caption{Quantitative results for each reconstruction method across clinical test-time datasets for (a) full-angle (FA) and (b) limited-angle (LA) acquisitions.}
\label{table:main_comparison}

\begin{tabularx}{\textwidth}{l l @{\;} l @{\extracolsep{\fill}} cYYcYYcYYcYY}
\toprule
& \multicolumn{2}{l}{Method} && \multicolumn{2}{c}{CERMEP} && \multicolumn{2}{c}{AGIEF} && \multicolumn{2}{c}{NX} && \multicolumn{2}{c}{Huntington} \\
\cmidrule{5-6} \cmidrule{8-9} \cmidrule{11-12} \cmidrule{14-15}
& &&& PSNR & SSIM && PSNR & SSIM && PSNR & SSIM && PSNR & SSIM \\
\midrule

\multirow{6}{*}{\rotatebox[origin=c]{90}{\textbf{a) FA}}} 
& OSEM        &       && 29.2 & 0.85 && 28.9 & 0.82 && 30.7 & 0.89 && 26.7 & 0.88 \\ 
& Bowsher     &       && 30.4 & 0.89 && 29.5 & 0.83 && 32.1 & 0.94 && \underline{28.0} & \underline{0.90} \\ 
& PET-PIR     & ($T=50$) && 24.5 & 0.84 && 27.2 & 0.77 && \underline{36.4} & \textbf{0.96} && 25.5 & 0.84 \\ 
& PET-DDS     & ($T=50$) && 27.4 & 0.83 && \underline{32.1} & \underline{0.91} && 35.7 & 0.94 && 24.9 & 0.81 \\ 
& LiSch-SCD   & ($T=50$) && \underline{32.7} & \underline{0.92} && 29.4 & 0.85 && 35.7 & \underline{0.95} && 28.3 & 0.87 \\ 
& Ours        & ($T=2$)  && \textbf{35.3} & \textbf{0.96} && \textbf{35.4} & \textbf{0.96} && \textbf{37.6} & \underline{0.95} && \textbf{30.7} & \textbf{0.91} \\ 

\midrule

\multirow{6}{*}{\rotatebox[origin=c]{90}{\textbf{b) LA}}} 
& OSEM        &       && 25.5 & 0.74 && 24.9 & 0.69 && 23.8 & 0.67 && 25.0 & \underline{0.75} \\ 
& Bowsher     &       && 26.2 & 0.77 && 25.6 & 0.71 && 25.7 & 0.75 && \textbf{25.9} & \textbf{0.78} \\ 
& PET-PIR     & ($T=50$) && 22.9 & 0.67 && 21.4 & 0.64 && 30.5 & \underline{0.90} && 22.7 & 0.69 \\ 
& PET-DDS     & ($T=50$) && 22.2 & 0.70 && \underline{28.4} & \underline{0.79} && 30.8 & 0.89 && 22.3 & 0.67 \\ 
& LiSch-SCD   & ($T=50$) && \underline{26.4} & \underline{0.81} && 27.3 & 0.78 && \underline{30.9} & \underline{0.90} && 23.4 & 0.70 \\ 
& Ours        & ($T=2$)  && \textbf{32.2} & \textbf{0.91} && \textbf{31.7} & \textbf{0.87} && \textbf{32.0} & \textbf{0.92} && \underline{25.2} & \textbf{0.78} \\ 
\bottomrule
\end{tabularx}
\end{table*}%

\noindent\textbf{Test-Time Domain Adaptation (TTDA):}
For PET image reconstruction, anatomical guidance from computed tomography (CT) or MR images is commonly used to regularize the reconstruction and improve image quality \cite{hashimoto2023fully,hashimoto2024deep}.
Given the superior soft-tissue contrast of magnetic resonance imaging (MRI) data, we focus on MRI-guided PET image reconstruction.
We condition the reconstruction model on MR images at test time by introducing a conditional branch that modulates the original network blocks via affine transformations \cite{perez2018film,stracke2024ctrloralter}.
To integrate this conditioning into the convolutional blocks of the U-Net-based pretrained score function $s_\theta$, we perform conditional low-rank adaptation \cite{stracke2024ctrloralter}.
Specifically, for a layer $l$ with input $\mathbf{x}^{l} \in \mathbb{R}^{k}$, the adapted output $\mathbf{h}^{l}$ is defined as follows:
\begin{equation}
\begin{array}{l}
\mathbf{h}^{l} = \mathbf{W}_0\mathbf{x}^{l} + \Delta\mathbf{W}\mathbf{x}^{l} 
= \mathbf{W}_0\mathbf{x}^{l} + \mathbf{U}\phi\left(\mathbf{D}\mathbf{x}^{l} \mid g(\mathbf{x}_\mathrm{MR})\right), \\
\text{with } \quad \phi(\mathbf{z} \mid g(\mathbf{x}_\mathrm{MR})) = \gamma_\phi(g(\mathbf{x}_\mathrm{MR})) \odot \mathbf{z} + \beta_\phi(g(\mathbf{x}_\mathrm{MR}))
  \end{array}
\end{equation}
Here, $\mathbf{W}_0, \Delta\mathbf{W} \in \mathbb{R}^{d \times k}$ denote the frozen pretrained weights and their adaptation, respectively.
The matrices $\mathbf{D} \in \mathbb{R}^{r \times k}$ and $\mathbf{U} \in \mathbb{R}^{d \times r}$ are learnable low-rank adaptation matrices with rank $r \ll \min(d,k)$.
The learnable mapper $g$ projects the conditioning input $\mathbf{x}_\mathrm{MR}$ to the latent adaptation space, while $\gamma_\phi$ and $\beta_\phi$ produce layer-wise scale and shift parameters from $g(\mathbf{x}_\mathrm{MR})$ (see Fig.~\ref{fig:main}b, c).
At test time, we adapt the pretrained model $s_\theta$ to target datasets by learning only these additional parameters.
We define an adaptation objective $L_\mathrm{adapt}$ combining a measurement-based loss ($L_\mathrm{p}$) and an MRI-driven anatomical loss ($L_\mathrm{a}$).
For $L_\mathrm{p}$, we minimize the Poisson NLL between the acquired measurements $\mathbf{y}$ and the forward projection $\hat{\mathbf{y}} = \mathbf{A}\hat{\mathbf{x}}^{\mathrm{MAP}} + \mathbf{b}$ \cite{10829716}.
The MRI-driven anatomical loss $L_\mathrm{a}$ is weighted by the total variation (TV) of $\mathbf{x}_\mathrm{MR}$, penalizing gradients more strongly in anatomically homogeneous regions and thereby encouraging the reconstruction to conform to the underlying anatomy.
From each MR image $\mathbf{x}_{\mathrm{MR}} \in \mathbb{R}^{n}_{\geq 0}$, we derive weights $w_p$, where $p \in \{0,1,...,n-1\}$, to encourage smoothing within homogeneous tissue regions while preserving edges aligned with anatomical boundaries.
We define
\begin{equation}
w_{j} = \exp\left(-\frac{\| (\nabla \mathbf{x}_{\text{MR}})_j \|}{\sigma}\right),
\end{equation} 
where $\| (\nabla \mathbf{x}_{\text{MR}})_j \|$ is the gradient magnitude at pixel $j$, and $\sigma$ controls sensitivity to anatomical edges.
The final adaptation objective is $L_\mathrm{adapt} = L_\mathrm{p} + \lambda L_\mathrm{a}$
where $\lambda$ controls the trade-off between data fidelity and anatomical regularization.
Unlike prior TTDA approaches that perform per-image adaptation \cite{hashimoto2026pet,webber2025steerable}, we perform dataset-level global adaptation for all reconstructions, reducing overhead while improving generalization and limiting per-image overfitting. \\
\noindent\textbf{Warm-starting (WS):}
Classical analysis of maximum-likelihood expectation-maxi\-mization (MLEM) or OSEM shows that low-frequency spatial components associated with large singular values of the system matrix converge rapidly, placing intermediate reconstructions close to the anatomical image manifold \cite{zeng2018maximum}.
Due to the ill-posed nature of the inverse problem, continued iterations tend to fit high-frequency Poisson noise rather than recover meaningful anatomical details, progressively driving the reconstruction away from the clean PET image manifold.
Guided by this geometric insight, we use the OSEM reconstructions to bypass the redundant low-frequency generation steps and later rely on the model $s_\theta$ to predict the high-frequency structures.
Specifically, we predict the noisy image corresponding to the OSEM images via diffusion inversion \cite{kim2022diffusionclip,songdenoising,yilmaz2025cellstyle} and initiate denoising from this noise representation instead of random noise (see Fig. \ref{fig:main}d).
This reduces the number of required diffusion steps substantially, without introducing additional computational overhead for computing the OSEM images, as these images are already required for measurement-based normalization regardless of the initialization strategy \cite{singh2023score,webber2025steerable} (see Section \ref{section:experiments}).
During inference, after WS, we perform $T$ alternating steps of prior refinements using $s_\theta$ and data-consistency updates to compute $\hat{\mathbf{x}}^{\mathrm{MAP}}$ (see Fig.~\ref{fig:main}e).
\section{Experiments and Results}\label{section:experiments}
\renewcommand{\floatpagefraction}{0.8}
\begin{figure}[!tbp]
\centering
\includegraphics[width=0.9\textwidth]{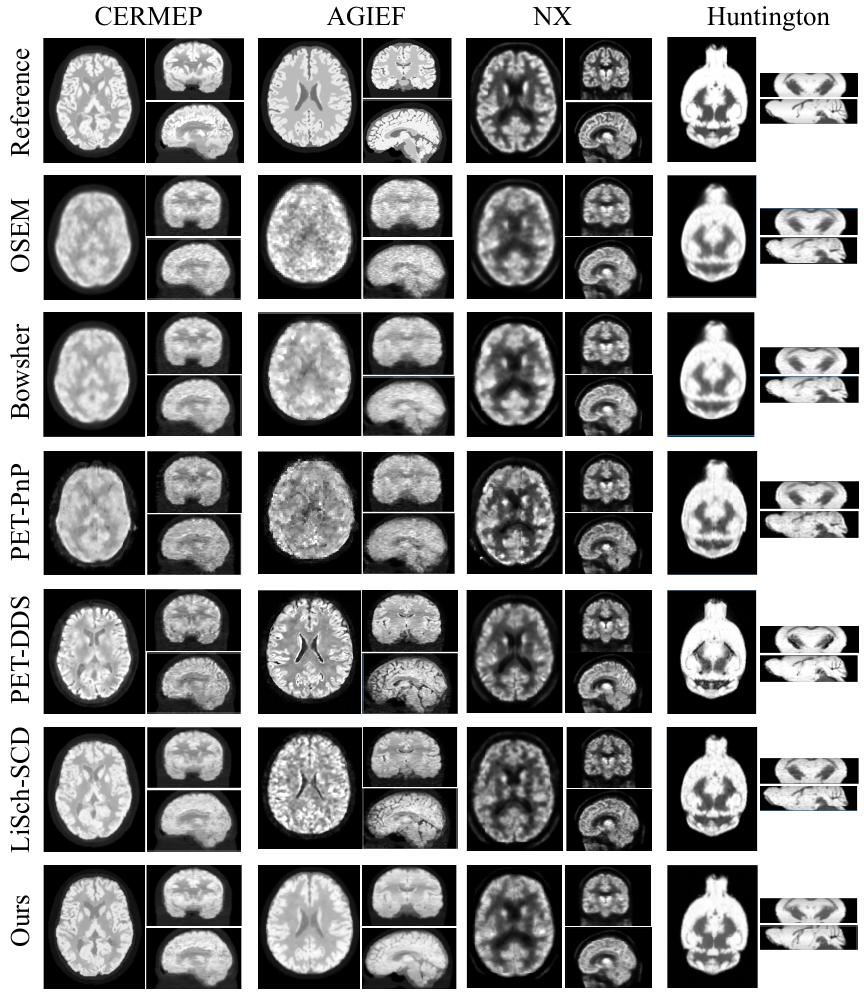}
\caption{Qualitative results for each reconstruction method and the reference GT volumes from axial (left), coronal (top right), and sagittal (bottom right) views.}
\label{fig:results}
\end{figure}%
\textbf{Datasets:} We work exclusively with joint PET–MRI datasets for anatomically guided PET image reconstruction, using brain phantom images for training and clinical datasets for domain adaptation experiments.
The BrainWeb dataset consists of 20 simulated FDG-PET volumes with coregistered MR images and represents a controlled, healthy human anatomy \cite{aubert2006twenty}.
CERMEP \cite{merida2021cermep} provides clinical brain PET–MRI data from healthy adults, which we use to evaluate the adaptation performance of the model trained on simulated phantoms on real data.
The AGIEF dataset \cite{zhang2022aerobic} includes scans from both healthy subjects and epilepsy patients, with unseen pathological metabolic patterns—such as focal hypometabolism and altered oxygen–glucose indices—not present in BrainWeb’s healthy simulations. 
The NeuroExplorer (NX) project provides multi-tracer data acquired using a long axial field-of-view PET/CT scanner, together with corresponding MR images, introducing previously unseen conditions arising from both novel hardware configurations and tracer distributions \cite{ds006917}.
To probe extreme anatomical domain shifts, we also use animal models of Huntington’s disease \cite{sawiak2016cambridge}.
Specifically, mouse brain MRI data are used to assign theoretical $[^{18}\text{F}]$FDG uptake values to segmented gray and white matter, thereby simulating reference PET images \cite{9123435}.
Unlike the NX dataset, the PET images provided in CERMEP and AGIEF are of low quality; therefore, for these datasets, we first generate high-SNR reference GT volumes by averaging the PET values within each brain region defined by the MRI segmentations. 
Next, for each dataset, we simulate PET measurements from the GT images using the Parallelproj framework \cite{schramm2024parallelproj}, modeling PET physics and noise, with true counts per emission volume set to 10 and 40\% of the detectors disabled during the LA simulations.
\begin{table*}[tbp]
\centering
\footnotesize
\newcolumntype{Y}{>{\centering\arraybackslash}X}
\setlength{\tabcolsep}{3pt} 

\caption{Ablation results comparing the full adaptation pipeline (PET-Adapter) to no warm-starting ($w/o$ WS) or no adaptation ($w/o$ TTDA) for (a) full-angle (FA) and (b) limited-angle (LA) acquisitions.}
\label{table:ablation}

\begin{tabularx}{\textwidth}{l l @{\quad} l @{\extracolsep{\fill}} cYYcYYcYYcYY}
\toprule
& \multicolumn{2}{l}{Method} && \multicolumn{2}{c}{CERMEP} && \multicolumn{2}{c}{AGIEF} && \multicolumn{2}{c}{NX} && \multicolumn{2}{c}{Huntington} \\
\cmidrule{5-6} \cmidrule{8-9} \cmidrule{11-12} \cmidrule{14-15}
& &&& PSNR & SSIM && PSNR & SSIM && PSNR & SSIM && PSNR & SSIM \\
\midrule

\multirow{4}{*}{\rotatebox[origin=c]{90}{\textbf{a) FA}}} 
& PET-Adapter   & ($T=2$)  && 35.3 & 0.96 && 35.4 & 0.96 && 37.6 & 0.95 && 30.7 & 0.91\\ 
& $w/o$ WS      & ($T=2$)       && 29.6 & 0.85 && 29.1 & 0.84 && 30.9 & 0.90 && 26.7 & 0.79 \\ 
& $w/o$ WS      & ($T=50$)       && 34.9 & 0.95 && 36.4 & 0.97 && 35.2 & 0.95 && 31.2 & 0.95 \\
& $w/o$ TTDA    & ($T=2$)       && 26.2 & 0.74 && 26.3 & 0.73 && 34.7 & 0.91 && 24.7 & 0.78 \\ 

\midrule

\multirow{4}{*}{\rotatebox[origin=c]{90}{\textbf{b) LA}}} 
& PET-Adapter & ($T=2$)  && 32.2 & 0.91 && 31.7 & 0.87 && 32.0 & 0.92 && 25.2 & 0.78 \\ 
& $w/o$ WS      & ($T=2$)       && 25.6 & 0.75 && 28.4 & 0.84 && 28.8 & 0.84 && 23.2 & 0.75 \\ 
& $w/o$ WS      & ($T=50$)       && 32.6 & 0.93 && 30.4 & 0.86 && 30.8 & 0.90 && 26.2 & 0.84 \\
& $w/o$ TTDA    & ($T=2$)       && 21.8 & 0.65 && 23.3 & 0.62 && 28.3 & 0.79 && 20.6 & 0.62 \\ 
\bottomrule
\end{tabularx}
\end{table*}\newline
\noindent\textbf{Training:} 
We train the score model $s_\theta$ on 2D slices of 3D images due to the limited number of training volumes.
$s_\theta$ is still used within a 3D reconstruction framework, where inter-slice consistency is enforced through warm-start initialization from 3D OSEM reconstructions. 
The same model weights are used for both FA and LA experiments, as the prediction of the data prior is independent of the measurement geometry.
The network is trained for 2000 epochs with a batch size of 32 and a learning rate of $1 \times 10^{-4}$ on an NVIDIA RTX 4000 GPU.
To account for large intensity variations across PET scans, image-wise normalization is applied to set the mean emission value of each training image to 1.0 \cite{singh2023score}. \\
\textbf{TTDA:} To adapt $s_\theta$ to out-of-distribution (OOD) data with additional anatomical conditioning, we set the hyperparameters based on two held-out samples from each target dataset \cite{10829716}.
For LA experiments, the contributions from missing sinogram regions are masked out when computing the loss $L_\mathrm{p}$. \\
\textbf{Inference:} During inference, we set the number of generation steps $T$ to 2, which is substantially fewer than in other generative PET reconstruction methods \cite{webber2025steerable,singh2023score}, enabled by warm-start initialization.
The prior predictions ($\hat{\mathbf{x}}_0$) from the score function $s_\theta$ are denormalized to the original image intensity range before the data consistency step to ensure a proper comparison between $\mathbf{y}$ and $\mathbf{\hat{y}}$ \cite{singh2023score}.
Since GT images are unavailable during adaptation and inference, the intensity range is estimated from OSEM reconstructions with $30$ subsets and $2$ iterations following \cite{singh2023score}. For warm-starting, we reverse the SDE to step $t_r=0.4$.\\
\noindent\textbf{Comparative Methods:} We evaluate our approach against established PET reconstruction methods using Peak Signal-to-Noise Ratio (PSNR) and Structural Similarity Index Measure (SSIM).
Baseline comparisons include classical algorithms, namely, OSEM \cite{osem} and Maximum a Posteriori Expectation Maximization with a Bowsher prior (MAPEM-Bowsher) \cite{bowsher2004utilizing}, where MR images are used to construct the anatomical prior.
We additionally compare against recent generative-model-based PET reconstruction methods.
These include PET-DDS \cite{singh2023score}, which applies classifier-free guidance (CFG) using MR images, and DiffPIR with feature injection conditioning \cite{tumanyan2023plug}, which we adapt to incorporate structural MRI information and PET-specific measurement updates, and rename as PET-PIR.
Finally, we compare to PET-LiSch-SCD (LiSch-SCD in short) \cite{webber2025steerable}, which applies the Steerable Conditional Diffusion framework \cite{10829716} to PET image reconstruction.
Following the published description, we implemented this method based on CFG-guided PET-DDS, as the original code is not publicly available.\\
\noindent\textbf{Results:} 
We report the quantitative results in terms of PSNR and SSIM in Table \ref{table:main_comparison}.
Notably, PETAdapter achieves successful TTDA using only $T=2$ generation steps, compared to the other diffusion-based methods listed in the table, which require $T=50$ steps.
As expected, the LA setting (\ref{table:main_comparison}b) yields lower quality than the FA setting (\ref{table:main_comparison}a) due to incomplete measurements.
Figure \ref{fig:results} presents qualitative comparisons in axial, coronal, and sagittal views for the FA setting (the LA reconstructions are not shown due to space constraints).
The visual results demonstrate that PETAdapter produces reconstructions with superior structural fidelity and intensity consistency compared to the other methods.\\
\noindent\textbf{Ablations:} 
To evaluate the contribution of different components in PETAdapter, we conduct ablation studies by omitting WS and TTDA.
In the absence of WS, test-time adaptation is still performed; however, the reconstruction is initialized from pure noise samples instead of OSEM reconstructions.
For this setting, we consider two configurations with $T=2$ and $T=50$ iterations.
With $T=2$, the reconstruction quality degrades substantially across all datasets compared to the full model with WS, as the limited number of iterations is insufficient for adequate prior refinement and measurement updates (see Table \ref{table:ablation}).
Increasing the number of iterations to $T=50$ restores the performance to a level comparable to $T=2$ with WS, highlighting the effectiveness of WS in reducing the required number of refinement steps.
To assess the impact of TTDA, we retain WS, set $T=2$, and omit adaptation to the target dataset.
This results in a significant performance drop due to domain shift, as shown in Table \ref{table:ablation}.
\section{Conclusion}
In this work, we introduced PET-Adapter for TTDA of generative PET image reconstruction models.
By enabling robust adaptation to novel clinical datasets, it improves the generalization across varying scanners, tracers, anatomies, and pathologies without extensive retraining.
Experiments in both FA and LA settings demonstrate that PET-Adapter achieves superior performance while reducing generative sampling steps from 50 to 2 via OSEM warm-starting.
Coupled with dataset-level adaptation, this reduction highlights its potential for immediate, computationally efficient clinical deployment.\\

\noindent\textbf{Acknowledgements:} We acknowledge the use of the CERMEP-iDB-MRXFDG dataset. © Copyright CERMEP -- Imagerie du vivant, www.cermep.fr and Hospices Civils de Lyon. All rights reserved.

\clearpage
\bibliographystyle{splncs04}
\bibliography{egbib}

@article{zhang2022aerobic,
  title={Aerobic glycolysis imaging of epileptic foci during the inter-ictal period},
  author={Zhang, Miao and Qin, Qikai and Zhang, Shuning and Liu, Wei and Meng, Hongping and Xu, Mengyang and Huang, Xinyun and Lin, Xiaozhu and Lin, Mu and Herman, Peter and others},
  journal={EBioMedicine},
  volume={79},
  year={2022},
  publisher={Elsevier}
}

@book{cherry2006pet,
  title={{PET}: Physics, Instrumentation, and Scanners},
  author={Cherry, Simon R and Dahlbom, Magnus},
  year={2006},
  publisher={Springer},
  editor={Phelps, Michael E}
}

@article{webber2024diffusion,
  title={Diffusion models for medical image reconstruction},
  author={Webber, George and Reader, Andrew J},
  journal={BJR| Artificial Intelligence},
  volume={1},
  number={1},
  pages={ubae013},
  year={2024},
  publisher={Oxford University Press}
}

@article{hashimoto2024deep,
  title={Deep learning-based {PET} image denoising and reconstruction: a review},
  author={Hashimoto, Fumio and Onishi, Yuya and Ote, Kibo and Tashima, Hideaki and Reader, Andrew J and Yamaya, Taiga},
  journal={Radiological Physics and Technology},
  volume={17},
  number={1},
  pages={24--46},
  year={2024},
  publisher={Springer}
}

@article{schaefferkoetter2020convolutional,
  title={Convolutional neural networks for improving image quality with noisy {PET} data},
  author={Schaefferkoetter, Josh and Yan, Jianhua and Ortega, Claudia and Sertic, Andrew and Lechtman, Eli and Eshet, Yael and Metser, Ur and Veit-Haibach, Patrick},
  journal={EJNMMI Research},
  volume={10},
  number={1},
  pages={105},
  year={2020},
  publisher={Springer}
}

@article{liu2019higher,
  title={Higher {SNR PET} image prediction using a deep learning model and {MRI} image},
  author={Liu, Chih-Chieh and Qi, Jinyi},
  journal={Physics in Medicine \& Biology},
  volume={64},
  number={11},
  pages={115004},
  year={2019},
  publisher={IOP Publishing}
}

@article{whiteley2020directpet,
  title={DirectPET: full-size neural network {PET} reconstruction from sinogram data},
  author={Whiteley, William and Luk, Wing K and Gregor, Jens},
  journal={Journal of Medical Imaging},
  volume={7},
  number={3},
  pages={032503--032503},
  year={2020},
  publisher={Society of Photo-Optical Instrumentation Engineers}
}

@article{liu2022deep,
  title={Deep-learning-based framework for {PET} image reconstruction from sinogram domain},
  author={Liu, Zhiyuan and Ye, Huihui and Liu, Huafeng},
  journal={Applied Sciences},
  volume={12},
  number={16},
  pages={8118},
  year={2022},
  publisher={MDPI}
}

@article{singh2023score,
  title={Score-Based Generative Models for {PET} Image Reconstruction},
  author={Singh, Imraj RD and Denker, Alexander and Barbano, Riccardo and Kereta, {\v{Z}}eljko and Jin, Bangti and Thielemans, Kris and Maass, Peter and Arridge, Simon and others},
  journal={Machine Learning for Biomedical Imaging},
  volume={2},
  pages={547--585},
  year={2024}
}

@ARTICLE{9123435,
  author={Mehranian, Abolfazl and Reader, Andrew J.},
  journal={IEEE Transactions on Radiation and Plasma Medical Sciences}, 
  title={Model-Based Deep Learning {PET} Image Reconstruction Using Forward–Backward Splitting Expectation–Maximization}, 
  year={2021},
  volume={5},
  number={1},
  pages={54-64}}

@article{hashimoto2023fully,
  title={Fully {3D} implementation of the end-to-end deep image prior-based {PET} image reconstruction using block iterative algorithm},
  author={Hashimoto, Fumio and Onishi, Yuya and Ote, Kibo and Tashima, Hideaki and Yamaya, Taiga},
  journal={Physics in Medicine \& Biology},
  volume={68},
  number={15},
  pages={155009},
  year={2023},
  publisher={IOP Publishing}
}

@ARTICLE{10025780,
  author={Ote, Kibo and Hashimoto, Fumio and Onishi, Yuya and Isobe, Takashi and Ouchi, Yasuomi},
  journal={IEEE Transactions on Medical Imaging}, 
  title={List-Mode {PET} Image Reconstruction Using Deep Image Prior}, 
  year={2023},
  volume={42},
  number={6},
  pages={1822-1834}}

@article{xie2024joint,
  title={Joint diffusion: mutual consistency-driven diffusion model for {PET-MRI} co-reconstruction},
  author={Xie, Taofeng and Cui, Zhuo-Xu and Luo, Chen and Wang, Huayu and Liu, Congcong and Zhang, Yuanzhi and Wang, Xuemei and Zhu, Yanjie and Chen, Guoqing and Liang, Dong and others},
  journal={Physics in Medicine \& Biology},
  volume={69},
  number={15},
  pages={155019},
  year={2024},
  publisher={IOP Publishing}
}

@article{raj2024recovery,
  title={Recovery of the spatially-variant deformations in dual-panel {PET} reconstructions using deep-learning},
  author={Raj, Juhi and Millardet, Ma{\"e}l and Krishnamoorthy, Srilalan and Karp, Joel S and Surti, Suleman and Matej, Samuel},
  journal={Physics in Medicine \& Biology},
  volume={69},
  number={5},
  pages={055028},
  year={2024},
  publisher={IOP Publishing}
}

@inproceedings{makkar2024partial,
  title={Partial-ring {PET} Image Correction using Implicit Neural Representation Learning},
  author={Makkar, Shubhangi and Ye, Siqi and B{\'e}guin, Marina and Dissertori, G{\"u}nther and Hrbacek, Jan and Lomax, Antony and McNamara, Keegan and Ritzer, Christian and Weber, Damien C and Xing, Lei and others},
  booktitle={20th International Conference on the use of Computers in Radiation Therapy, Lyon, France},
  pages={8--11},
  year={2024}
}

@INPROCEEDINGS{11287858,
  author={Yilmaz, R. and Thull, J. and Stegmaier, J. and Schulz, V.},
  booktitle={2025 IEEE Nuclear Science Symposium (NSS), Medical Imaging Conference (MIC) and Room Temperature Semiconductor Detector Conference (RTSD)}, 
  title={Diffusion-Based Sinogram Interpolation for Limited Angle {PET}}, 
  year={2025},
  volume={},
  number={},
  pages={1-1}}

@inproceedings{stracke2024ctrloralter,
  title={{CTRLorALTer}: conditional loradapter for efficient 0-shot control and altering of t2i models},
  author={Stracke, Nick and Baumann, Stefan Andreas and Susskind, Joshua and Bautista, Miguel Angel and Ommer, Bj{\"o}rn},
  booktitle={European Conference on Computer Vision},
  pages={87--103},
  year={2024},
  organization={Springer}
}

@inproceedings{perez2018film,
  title={Film: Visual reasoning with a general conditioning layer},
  author={Perez, Ethan and Strub, Florian and De Vries, Harm and Dumoulin, Vincent and Courville, Aaron},
  booktitle={Proceedings of the AAAI Conference on Artificial Intelligence},
  volume={32},
  number={1},
  year={2018}
}

@article{10829716,
  title={Steerable conditional diffusion for out-of-distribution adaptation in medical image reconstruction},
  author={Barbano, Riccardo and Denker, Alexander and Chung, Hyungjin and Roh, Tae Hoon and Arridge, Simon and Maass, Peter and Jin, Bangti and Ye, Jong Chul},
  journal={IEEE Transactions on Medical Imaging},
  volume={44},
  number={5},
  pages={2093--2104},
  year={2025},
  publisher={IEEE}
}

@article{zeng2018maximum,
  title={Maximum-likelihood expectation-maximization algorithm versus windowed filtered backprojection algorithm: a case study},
  author={Zeng, Gengsheng L},
  journal={Journal of Nuclear Medicine Technology},
  volume={46},
  number={2},
  pages={129--132},
  year={2018},
  publisher={Society of Nuclear Medicine}
}

@inproceedings{webber2025steerable,
  title={Steerable Conditional Diffusion for Domain Adaptation in {PET} Image Reconstruction},
  author={Webber, G and Hammers, A and King, AP and Reader, AJ},
  booktitle={2025 IEEE Nuclear Science Symposium (NSS), Medical Imaging Conference (MIC) and Room Temperature Semiconductor Detector Conference (RTSD)},
  pages={1--1},
  year={2025},
  organization={IEEE}
}

@article{aubert2006twenty,
  title={Twenty new digital brain phantoms for creation of validation image data bases},
  author={Aubert-Broche, Bereng{\`e}re and Griffin, Mark and Pike, G Bruce and Evans, Alan C and Collins, D Louis},
  journal={IEEE Transactions on Medical Imaging},
  volume={25},
  number={11},
  pages={1410--1416},
  year={2006},
  publisher={IEEE}
}

@article{merida2021cermep,
  title={{CERMEP-IDB-MRXFDG}: a database of 37 normal adult human brain {[18F] FDG PET, T1 and FLAIR MRI, and CT} images available for research},
  author={M{\'e}rida, In{\'e}s and Jung, Julien and Bouvard, Sandrine and Le Bars, Didier and Lancelot, Sophie and Lavenne, Franck and Bouillot, Caroline and Redout{\'e}, J{\'e}r{\^o}me and Hammers, Alexander and Costes, Nicolas},
  journal={EJNMMI Research},
  volume={11},
  number={1},
  pages={91},
  year={2021},
  publisher={Springer}
}

@article{ds006917,
  title={Exceptional brain {PET} images from the {NeuroEXPLORER}: scans with targeted radiopharmaceuticals and comparison to {HRRT}},
  author={Volpi, Tommaso and Toyonaga, Takuya and Khattar, Nikkita and Gallezot, Jean-Dominique and Naganawa, Mika and Vanderlinden, Greet and Honhar, Praveen and Zeng, Tianyi and Fontaine, Kathryn and Mulnix, Tim and others},
  journal={European Journal of Nuclear Medicine and Molecular Imaging},
  pages={1--12},
  year={2025},
  publisher={Springer}
}

@article{sawiak2016cambridge,
  title={The Cambridge {MRI} database for animal models of Huntington disease},
  author={Sawiak, Stephen J and Morton, A Jennifer},
  journal={NeuroImage},
  volume={124},
  pages={1260--1262},
  year={2016},
  publisher={Elsevier}
}

@article{schramm2024parallelproj,
  title={PARALLELPROJ—an open-source framework for fast calculation of projections in tomography},
  author={Schramm, Georg and Thielemans, Kris},
  journal={Frontiers in Nuclear Medicine},
  volume={3},
  pages={1324562},
  year={2024},
  publisher={Frontiers Media SA}
}

@inproceedings{tumanyan2023plug,
  title={Plug-and-play diffusion features for text-driven image-to-image translation},
  author={Tumanyan, Narek and Geyer, Michal and Bagon, Shai and Dekel, Tali},
  booktitle={Proceedings of the IEEE/CVF Conference on Computer Vision and Pattern Recognition},
  pages={1921--1930},
  year={2023}
}

@article{vincent2011connection,
  title={A Connection Between Score Matching and Denoising Autoencoders},
  author={Vincent, Pascal},
  journal={Neural Computation},
  volume={23},
  number={7},
  pages={1661--1674},
  year={2011},
  publisher={MIT Press}
}

@inproceedings{songscore,
  title={Score-Based Generative Modeling through Stochastic Differential Equations},
  author={Song, Yang and Sohl-Dickstein, Jascha and Kingma, Diederik P and Kumar, Abhishek and Ermon, Stefano and Poole, Ben},
  booktitle={International Conference on Learning Representations},
  year=2021
}

@article{osem,
  title={Accelerated image reconstruction using ordered subsets of projection data},
  author={Hudson, H Malcolm and Larkin, Richard S},
  journal={IEEE Transactions on Medical Imaging},
  volume={13},
  number={4},
  pages={601--609},
  year={1994},
  publisher={IEEE}
}

@inproceedings{bowsher2004utilizing,
  title={Utilizing {MRI} information to estimate {F18-FDG} distributions in rat flank tumors},
  author={Bowsher, James E and Yuan, Hong and Hedlund, Laurence W and Turkington, Timothy G and Akabani, Gamal and Badea, Alexandra and Kurylo, William C and Wheeler, C Ted and Cofer, Gary P and Dewhirst, Mark W and others},
  booktitle={IEEE Symposium Conference Record Nuclear Science 2004.},
  volume={4},
  pages={2488--2492},
  year={2004},
}

@inproceedings{songdenoising,
  title={Denoising Diffusion Implicit Models},
  author={Song, Jiaming and Meng, Chenlin and Ermon, Stefano},
  booktitle={International Conference on Learning Representations},
  year=2021
}

@article{hashimoto2026pet,
  title={Pet image reconstruction using deep diffusion image prior},
  author={Hashimoto, Fumio and Gong, Kuang},
  journal={IEEE Transactions on Medical Imaging},
  year={2026},
  publisher={IEEE}
}

@inproceedings{kim2022diffusionclip,
  title={Diffusionclip: Text-guided diffusion models for robust image manipulation},
  author={Kim, Gwanghyun and Kwon, Taesung and Ye, Jong Chul},
  booktitle={Proceedings of the IEEE/CVF conference on computer vision and pattern recognition},
  pages={2426--2435},
  year={2022}
}

@inproceedings{yilmaz2025cellstyle,
  title={CellStyle: Improved Zero-Shot Cell Segmentation via Style Transfer},
  author={Yilmaz, R{\"u}veyda and Chen, Zhu and Wu, Yuli and Stegmaier, Johannes},
  booktitle={International Conference on Medical Image Computing and Computer-Assisted Intervention},
  pages={67--77},
  year={2025},
  organization={Springer}
}
\end{document}